\documentclass{ecai} 
\usepackage{latexsym}
\usepackage{amssymb}
\usepackage{amsmath}
\usepackage{amsthm}
\usepackage{booktabs}
\usepackage{enumitem}
\usepackage{graphicx}
\usepackage{color}
\usepackage{multirow}

\newcommand{\BibTeX}{B\kern-.05em{\sc i\kern-.025em b}\kern-.08em\TeX}

\begin{document}

%%%%%%%%%%%%%%%%%%%%%%%%%%%%%%%%%%%%%%%%%%%%%%%%%%%%%%%%%%%%%%%%%%%%%%%%

\begin{frontmatter}

%%% Use this command to specify your submission number.
%%% In doubleblind mode, it will be printed on the first page.

\paperid{123} 

%%% Use this command to specify the title of your paper.

\title{CCRC: A Change-Aware Captioning and Reasoning Chain for Image Change Captioning and Segmentation}

%%% Use this combinations of commands to specify all authors of your 
%%% paper. Use \fnms{} and \snm{} to indicate everyone's first names 
%%% and surname. This will help the publisher with indexing the 
%%% proceedings. Please use a reasonable approximation in case your 
%%% name does not neatly split into "first names" and "surname".
%%% Specifying your ORCID digital identifier is optional. 
%%% Use the \thanks{} command to indicate one or more corresponding 
%%% authors and their email address(es). If so desired, you can specify
%%% author contributions using the \footnote{} command.

\author[A]{\fnms{Jinhong}~\snm{Hu}}
\author[A]{\fnms{Xiaoping}~\snm{Wang}\thanks{Corresponding Authors. Email: xpwang@hnu.edu.cn, kailu@nudt.edu.cn}}
\author[B]{\fnms{Shuyin}~\snm{Huang}}
\author[A]{\fnms{Guojin}~\snm{Zhong}}
\author[A]{\fnms{Kaitai}~\snm{Liu}}
\author[C]{\fnms{Kai}~\snm{Lu}\footnotemark[*]}

\address[A]{College of Computer Science and Electronic Engineering, Hunan University, China}
\address[B]{Guangxi Minzu University, China}
\address[C]{National University of Defense Technology, China}

%%% Use this environment to include an abstract of your paper.

\begin{abstract}
Understanding and localizing subtle changes between paired images is critical for tasks such as surveillance and image editing. However, traditional Image Change Captioning (ICC) methods lack spatial grounding, limiting their precision. We introduce Image Change Captioning and Segmentation (ICCS), a new multimodal task that jointly requires structured change description and pixel-level localization. To address ICCS, we propose the Change-aware Captioning and Reasoning Chain (CCRC), a dual-chain framework that decouples semantic reasoning from spatial segmentation. The first chain, Chain-of-Change-Captioning (CCC), enhances fine-grained change perception via a visual fusion module—based on Multi-Head Change-aware Attention—inserted between the visual and language components of a Multimodal Large Language Model (MLLM). CCC also determines whether a change is segmentable. If not, it alone generates the caption. Otherwise, the second chain, Chain-of-Change-Segmenting (CCS), is activated, leveraging spatial priors from CCC and refining masks with a Change-aware Token Refiner for accurate boundary localization. We evaluate CCRC on both synthetic and real-world change detection benchmarks with pixel-level supervision. Experiments show CCRC achieves state-of-the-art performance. Code is available at \url{https://github.com/user-jinhong/CCRC}.
\end{abstract}

\end{frontmatter}

%%%%%%%%%%%%%%%%%%%%%%%%%%%%%%%%%%%%%%%%%%%%%%%%%%%%%%%%%%%%%%%%%%%%%%%%

\section{Introduction}
\begin{figure}[htbp]
    \centering
    \includegraphics[width=0.9\linewidth]{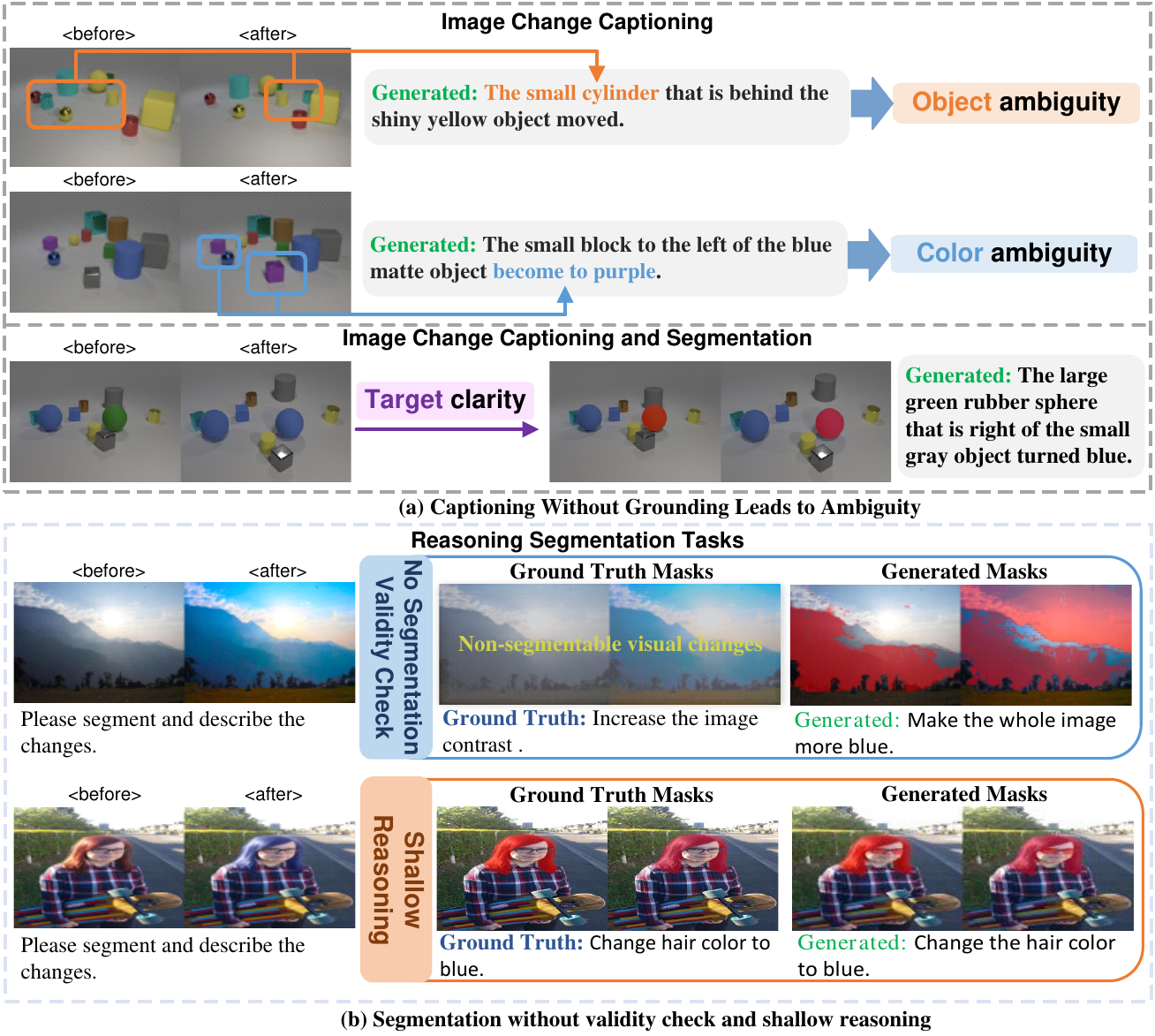}
    \caption{Illustration of Ambiguity and Failure Cases in Existing ICC and MLLM-based Segmentation Methods.}
    \label{fig:overview}
\end{figure}

Understanding and localizing visual changes between image pairs is a key challenge in vision-language reasoning. Traditional Image Change Captioning (ICC) models \cite{park2019robust} focus on generating descriptions but are limited in real-world applications due to their lack of spatial awareness. A fundamental issue with ICC is the inherent \textbf{semantic ambiguity} of the generated captions. For example, as shown in Figure~\ref{fig:overview}(a), the description \textit{“The small cylinder that is behind the shiny yellow object moved”} leads to \textbf{object ambiguity}, as multiple small cylinders may exist within the scene. Similarly, the description \textit{“The small block to the left of the blue matte object turned purple”} reflects \textbf{color ambiguity}, since several objects may share similar purplish appearances. The lack of explicit spatial grounding and validity verification results in descriptions that are insufficiently discriminative for precise localization, highlighting the need to integrate structured reasoning and spatial disambiguation into change captioning tasks.

To address these issues, we propose a new multimodal task: Image Change Captioning and Segmentation (ICCS). In this task, the model generates a natural language description of the change and a precise segmentation mask of the changed region, enabling joint semantic and spatial understanding. This is crucial for real-world applications such as surveillance, image editing, and ecological monitoring. As shown in Figure~\ref{fig:overview}(a), ICCS directly aligns the caption (e.g., "The large green rubber sphere to the right of the small gray object turned blue") with the visualized change region, significantly enhancing the discriminative clarity of the expression and effectively eliminating the localization ambiguities that are inherent in traditional ICC, which relies solely on language.

Although ICCS unifies semantic and spatial reasoning, most methods \cite{xu2023bridging} \cite{xia2024gsva} still separate captioning and segmentation, lacking an integrated reasoning framework for real-world change detection and localization. With the rapid advancement of Multimodal Large Language Models (MLLMs), it is now feasible to build end-to-end systems that unify semantic and spatial reasoning through coherent interaction. Representative attempts, such as LISA~\cite{lai2024lisa}, combine LLaVA \cite{liu2024visual} with a SAM decoder \cite{kirillov2023segment} and use synthetic supervision to enable language-guided segmentation. However, as illustrated in Figure~\ref{fig:overview}(b), such approaches still face three key challenges when applied to the ICCS setting: (1) \textbf{Lack of segmentation validity check.} The model fails to recognize non-localizable changes. For example, in the “overall brightness increase” case, which reflects only a global style adjustment, it still generates incorrect partial masks. (2) \textbf{Shallow reasoning.} In fine-grained changes like “hair color modification,” the model fails to focus on the key region, producing masks that not only miss the target but also include irrelevant areas. (3) \textbf{Lack cross-image contrast modeling.} Most MLLM-based methods process the image pair independently or via naive concatenation, without explicitly capturing their differences, making it difficult to detect subtle movements or attribute shifts.

To address the limitations of existing MLLM-based approaches—such as weak spatial discrimination, insensitivity to fine-grained changes, and inability to handle non-segmentable cases—we propose a dual-chain reasoning framework called \textbf{CCRC (Change-aware Captioning and Reasoning Chain)}. Built on a pretrained vision-language model (e.g., LLaVA), CCRC decouples semantic reasoning from spatial localization, while preserving reasoning continuity via a sequential design. The first chain, \textit{Chain-of-Change-Captioning (CCC)}, performs image-text alignment using the MLLM backbone and enhances inter-image difference modeling through a \textbf{Visual Fusion} module, where a \textit{Multi-Head Change-Aware Attention} mechanism is introduced to strengthen contrastive reasoning. This mechanism improves the model's ability to distinguish subtle differences between images while suppressing irrelevant background noise. To further improve change localization and description, we introduce an \textbf{In-Context Guidance} that injects spatial and semantic priors into the input context. CCC then generates a structured change description and predicts a segmentation flag token \textbf{[SF]} to determine whether spatial localization is needed. If the change is deemed segmentable, the second chain, \textit{Chain-of-Change-Segmenting (CCS)}, is triggered. CCS performs coarse localization based on spatial priors from CCC and progressively refines the change mask using a \textit{Change-Aware Token Refiner (CATRefiner)} for pixel-level precision. If the change is judged as non-segmentable (e.g., global style shift or no significant visual difference), CCC alone generates the textual description, and the segmentation process is skipped. 

The key contributions of this work are as follows:
\begin{itemize}
    \item We define a new vision-language task, \textbf{Image Change Captioning and Segmentation (ICCS)}, which jointly performs change description and pixel-level localization.
    \item We propose \textbf{CCRC}, a dual-chain architecture that decouples semantic reasoning from spatial localization. The first chain, \textit{CCC}, generates structured change descriptions through visual fusion enhanced by multi-head change attention and in-context guidance. The second chain, \textit{CCS}, uses CCC-derived spatial priors for coarse localization and refines segmentation masks via a CATRefiner.
    \item We extend two existing benchmarks, CLEVR-Change and Image-Editing-Request, with fine-grained segmentation annotations to support ICCS evaluation, and demonstrate that CCRC achieves state-of-the-art performance on both captioning and segmentation tasks.
\end{itemize}

\section{Related Work}
\subsection{Image Change Captioning}
Image Change Captioning (ICC) focuses on describing subtle visual differences between similar images. Early works like DUDA \cite{park2019robust} and VAM \cite{shi2020finding} enhance change localization via attention and viewpoint modeling. Various subsequent models have continued to refine change localization and semantic alignment \cite{tu2021rˆ3net, tu2024context,tu2024distractors, huang2021image}. IDC-PCL \cite{yao2022image} introduces contrastive learning and self-supervised alignment of vision-language features. Recent transformer-based models \cite{qiu2021describing,tu2023adaptive}, improve region-word correspondence. \cite{hu2025mct, zhong2025decider} introduce diffusion models \cite{ho2020denoising} to enhance change representation and generate more accurate captions in ICC. FINER-MLLM \cite{zhang2024differential} further leverages multimodal large language models (MLLMs) with dual feature constraints and retrieval-based enhancement for fine-grained captioning.

We propose \textbf{Image Change Captioning and Segmentation (ICCS)}, a task that combines textual generation and pixel-level localization, extending the capabilities of traditional ICC. Unlike models that produce vague descriptions, ICCS enables more accurate reasoning and segmentation.

\subsection{Multimodal Large Language Model}
Multimodal Large Language Models (MLLMs) integrate visual and textual inputs to enhance performance in vision-language tasks. Representative models such as  LLaVA \cite{liu2024visual} leverage frozen vision encoders with cross-modal fusion or fine-tuning to enable effective in-context learning. InstructBLIP \cite{dai2023instructblip} further improves instruction-following via curated datasets. Beyond captioning, MLLMs have been extended to segmentation tasks, such as LISA \cite{lai2024lisa} and CoRes~\cite{bao2024cores}, which incorporate reasoning-based segmentation. However, these models struggle to capture subtle image-level Changes. FINER-MLLM \cite{zhang2024differential} is the first to introduce MLLMs for image change captioning, using feature-level constraints for change perception. Nevertheless, it lacks segmentation capability.
Our method extends MLLMs to jointly perform change captioning and segmentation, enabling spatially grounded and semantically precise understanding of visual Changes.

\subsection{Image Segmentation}
Referring Image Segmentation (RIS) segments targets from natural language queries, with early works combining CNN \cite{hu2016segmentation}, and later models using attention and Transformers. SAM \cite{kirillov2023segment} enhances segmentation quality by using high-quality masks, and Transformer models \cite{tang2023contrastive} and ETRIS \cite{xu2023bridging} improve fusion and optimization through cross-modal decoders. Reasoning segmentation \cite{lai2024lisa} \cite{bao2024cores} predicts binary masks from image-text pairs but lacks the ability to model changes between images. In contrast, ICCS aims to capture subtle, context-dependent Changes across image pairs. Unlike semantic or instance segmentation, it focuses on dynamic variations rather than static object labeling. This makes ICCS more challenging, requiring both precise localization and change-aware reasoning.

\begin{figure*}[t]
    \centering
    \includegraphics[width=0.88\textwidth]{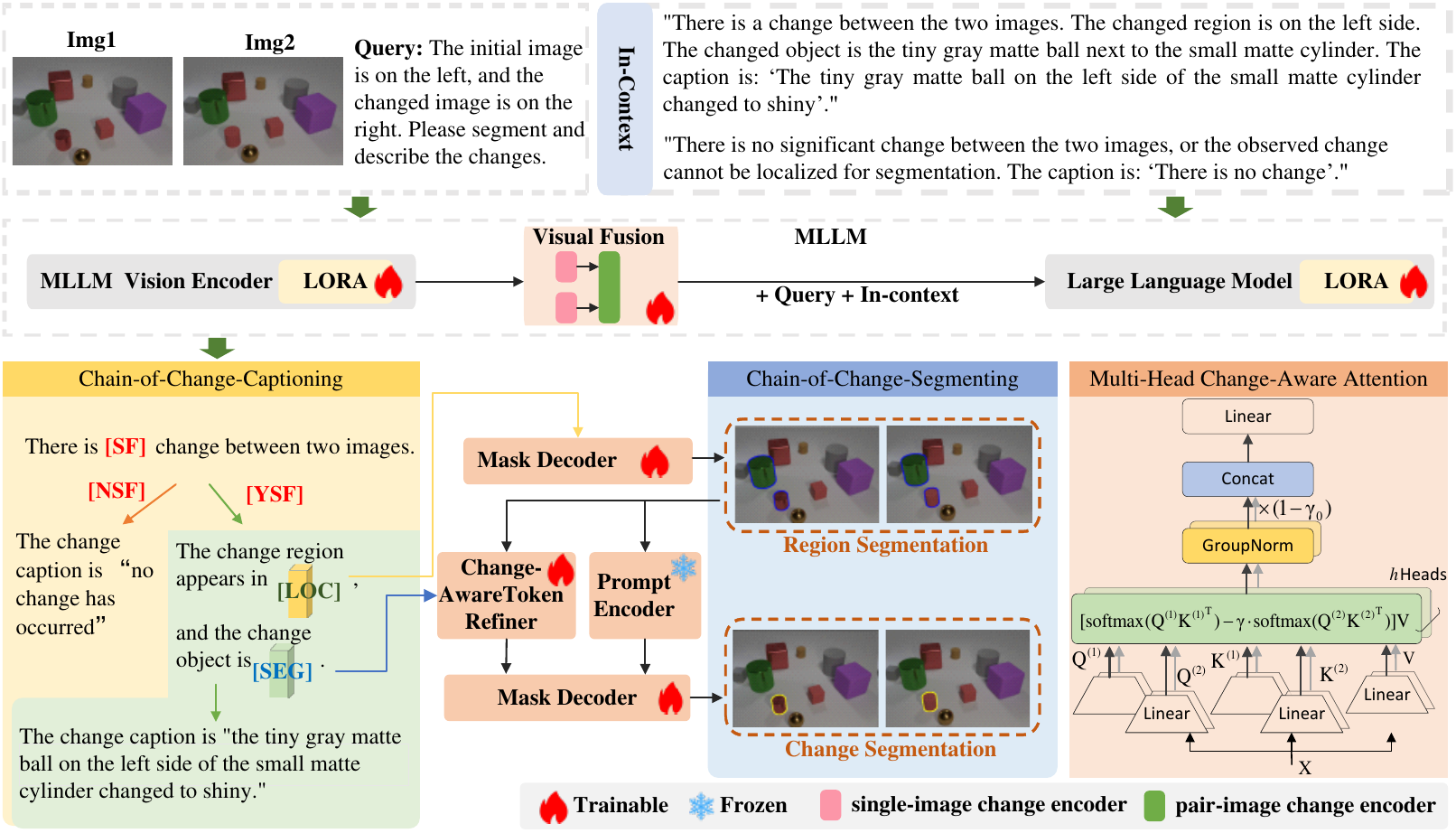}
\caption{
Overview of the proposed CCRC framework for the ICCS task. Given a pair of aligned images, visual features are extracted by the MLLM (SAM encoder omitted for clarity), enhanced via a Visual Fusion module based on Multi-Head Change-Aware Attention, and combined with a change-specific in-context prompt. These fused features and prompts are processed by a Large Language Model (LLM), forming the core of the Chain-of-Change-Captioning (CCC) module, which predicts a segmentation flag \textbf{[SF]}, spatial prompts \textbf{[LOC]} and \textbf{[SEG]}, and generates a structured caption. If \textbf{[SF] = [NSF]}, indicating a non-segmentable change, only the caption is produced. Otherwise, if \textbf{[SF] = [YSF]}, the Chain-of-Change-Segmenting (CCS) module is activated. CCS first performs \textit{Region Segmentation} based on the spatial cues provided by CCC (shown in blue), and then refines the segmentation within the region through the Change-Aware Token Refiner (CATRefiner) to produce fine-grained \textit{Change Segmentation} (shown in red).
}
    \label{fig:ccrc_overview}
\end{figure*}

\section{Task Definition}
Existing Image Change Captioning (ICC) tasks often suffer from semantic ambiguity due to the lack of spatial grounding, making it hard to determine what changed and where. To address this, we propose \textbf{Image Change Captioning and Segmentation (ICCS)}, a new task that jointly detects, describes, and segments semantic changes between aligned image pairs. Given two input images, the model determines whether a change occurred, generates a structured caption indicating its content and location, and segments the changed region at pixel level. ICCS unifies semantic description and spatial localization, bridging the gap between conventional ICC—which lacks grounding—and referring segmentation—which focuses on single-object queries. Unlike existing MLLMs that struggle with contrastive reasoning and segmentability prediction, ICCS provides a more structured benchmark for grounded multimodal understanding.

\section{Methodology}

\subsection{Architecture Overview}

We propose the Change-aware Captioning and Reasoning Chain (\textbf{CCRC}), an end-to-end model tailored for the Image Change Captioning and Segmentation (ICCS) task. At its core, CCRC adopts a dual-chain architecture that decouples semantic reasoning from spatial localization. The first chain generates structured change descriptions, enhanced by a lightweight change-aware attention module that strengthens cross-image contrast. The second chain then performs fine-grained segmentation guided by the spatial cues from the first.

As illustrated in Figure~\ref{fig:ccrc_overview}, CCRC adopts a progressive dual-chain architecture that enables adaptive switching between semantic-level inference and detailed spatial localization. By explicitly decoupling what has changed from where it occurs, the model provides a unified and interpretable solution to image change comprehension. This dual-chain reasoning process begins with the \textit{Chain-of-Change-Captioning (CCC)} (see Sec.~\ref{sec:ccc}), which takes contrast-enhanced visual features and in-context prompts as input and generates a structured output beginning with a segmentation flag token \(\textbf{[SF]}\), followed by a natural language description \(T\). If \(\textbf{[SF]} = \textbf{[NSF]}\), indicating the change is not segmentable, the pipeline terminates after caption generation. Here, non-segmentable changes refer to cases where the Change cannot be localized to a specific region, such as no visual change or global image adjustments like brightness or filter effects. Otherwise, when \(\textbf{[SF]} = \textbf{[YSF]}\), CCC additionally predicts a spatial prior \(\textbf{[LOC]}_i\) and a coarse segmentation token \(\textbf{[SEG]}_i\), which are passed to the \textit{Chain-of-Change-Segmenting (CCS)} (see Sec.~\ref{sec:CCS}). Rather than performing segmentation over the entire image, CCS first uses \(\textbf{[LOC]}_i\) to coarsely localize the region of change. Guided by this region-level cue, the model then refines the segmentation within the localized area to produce pixel-accurate masks \(\textbf{[SEG]}_i\) that precisely delineate the changed object. The coarse-to-fine strategy improves segmentation accuracy while reducing background interference. Through the integration of semantic reasoning and spatial grounding, CCRC provides a unified framework that captures both what changed and where it occurred, enabling precise and interpretable visual change understanding.

\subsection{Chain-of-Change-Captioning}
\label{sec:ccc}
The \textit{Chain-of-Change-Captioning (CCC)} serves as the semantic reasoning stage in the CCRC model. It is designed to generate structured natural language descriptions that explicitly articulate the semantic Changes between a pair of input images. Unlike traditional image captioning tasks that focus on static understanding of a single image, CCC formulates change comprehension as a contrastive reasoning process conditioned on paired visual features and task-specific language prompts. Specifically, CCC is built upon a Multimodal Large Language Model (MLLM), leveraging its strong cross-modal alignment capability to perform joint visual-linguistic modeling for high-quality change description. This reasoning chain consists of three core components: a visual fusion module for contrast modeling, an in-context prompting mechanism for language conditioning, and a captioning head for structured output generation.

% --------------------------
\textbf{Visual Fusion.}
To enable fine-grained reasoning over image Changes, the visual fusion module adopts a two-stage architecture consisting of a \textit{single-image change encoder} and a \textit{pair-image change encoder}. This architecture explicitly decouples intra-image context modeling from inter-image contrast integration, allowing the model to progressively capture both structural context and change cues.

Given visual features \( F_1, F_2 \in \mathbb{R}^{N \times d} \) extracted by a shared-weight MLLM visual encoder, each image is first independently processed by an intra-image change encoder \( \mathcal{A}_{\text{intra}}^{\Delta}(\cdot) \) to preserve its contextual representation. The resulting features are then concatenated and fed into an inter-image change encoder \( \mathcal{A}_{\text{inter}}^{\Delta}(\cdot) \), which explicitly models the contrast between the two images:
\begin{equation}
F_{\text{fused}} = \mathcal{A}_{\text{inter}}^{\Delta}\left(
\left[
\mathcal{A}_{\text{intra}}^{\Delta}(F_1); \ 
\mathcal{A}_{\text{intra}}^{\Delta}(F_2)
\right]
\right).
\end{equation}
The contrast-enhanced representation \( F_{\text{fused}} \) is then passed to the downstream Chain-of-Change-Captioning (CCC) module for structured reasoning.

Both $\mathcal{A}_{\text{self}}^{\Delta}(\cdot)$ and $\mathcal{A}_{\text{cross}}^{\Delta}(\cdot)$ employ multi-head change-aware attention to amplify token-level changes and suppress redundant or shared attention noise (see Figure~\ref{fig:ccrc_overview}). 
While the subtraction-based formulation resembles that of~\cite{cang2025shared}, our design focuses on cross-modal contrast and spatial change localization in paired visual inputs by applying change-aware attention to both self- and cross-attention pathways.

For each attention head, the input feature matrix \( X \in \mathbb{R}^{N \times d_{\text{model}}} \) is projected as:
\begin{equation}
[Q^{(1)}, Q^{(2)}] = X W_Q, \quad [K^{(1)}, K^{(2)}] = X W_K, \quad V = X W_V
\end{equation}
where \( W_Q, W_K \in \mathbb{R}^{d_{\text{model}} \times 2d} \) and \( W_V \in \mathbb{R}^{d_{\text{model}} \times d} \) are learnable parameters. The attention map is computed via subtraction of dual attention scores:
\begin{equation}
\mathbf{M}^{\Delta} = \text{Softmax}\left(\frac{Q^{(1)} {K^{(1)}}^\top}{\sqrt{d}}\right) 
- \gamma \cdot \text{Softmax}\left(\frac{Q^{(2)} {K^{(2)}}^\top}{\sqrt{d}}\right)
\end{equation}
where \( \gamma \) is a learnable contrast-suppression factor, reparameterized as:
\begin{equation}
\gamma = \exp(u_1^\top v_1) - \exp(u_2^\top v_2) + \gamma_0
\end{equation}
with \( u_1, u_2, v_1, v_2 \in \mathbb{R}^d \), which are learnable parameters, and \( \gamma_0 \) a fixed scalar set to 0.8 for all layers.

Each attention head then computes the output via:
\begin{equation}
\text{head}_i = (1 - \gamma_0) \cdot \text{Norm}(\mathbf{M}^{\Delta} V^{(i)})
\end{equation}
where \(\text{Norm}(\cdot)\) denotes per-head normalization (e.g., LayerNorm or GroupNorm). Multi-head outputs are concatenated and linearly projected:
\begin{equation}
\text{MultiHead}(X) = \text{Concat}(\text{head}_1, \dots, \text{head}_h) W^O
\end{equation}
with \( W^O \in \mathbb{R}^{hd \times d_{\text{model}}} \) as the final output projection.

The change attention module is integrated into a residual Transformer block with SwiGLU-activated feed-forward layers, enabling stable learning and effective feature composition.

The Visual Fusion mechanism, with its two-stage architecture and change-aware attention, captures subtle image differences and localizes changes effectively. By decoupling intra-image context and inter-image contrast, it enhances fine-grained reasoning and spatial localization for precise change detection and segmentation.

% --------------------------
\textbf{In-Context Guidance.}  
To enable structured generation, CCC adopts an in-context prompting strategy during training, where each prompt consists of two natural language exemplars—one with a segmentable change and one with a non-segmentable case. These exemplars follow a fixed linguistic template: the structural description is generated by GPT \cite{brown2020language}, while the caption comes directly from the dataset’s ground-truth annotation.

As shown in the in-context exemplars in the figure \ref{fig:ccrc_overview}, the segmentable example demonstrates a well-structured in-context prompt composed of four natural language sentences. The first sentence, \textit{"There is a change between the two images."}, expresses the presence of a visual Change, implicitly guiding the model to predict a positive segmentation flag (\textbf{[SF]} = \textbf{[YSF]}). The second sentence, \textit{"The changed region is on the left side."}, provides a coarse spatial cue, functioning as the prior for \textbf{[LOC]}. The third sentence, \textit{"The changed object is the tiny gray matte ball next to the small matte cylinder."}, specifies the target entity of the change, corresponding to \textbf{[SEG]}. Finally, the fourth sentence, beginning with \textit{"The caption is:"}, delivers the natural language description that serves as the model's generated caption \textbf{[T]}. In contrast, the non-segmentable example lacks such structured spatial and object-level references. Its first sentence, \textit{"There is no significant change between the two images, or the observed change cannot be localized for segmentation."}, conveys either the absence of meaningful change or the presence of a global transformation. This leads the model to infer a negative segmentation flag (\textbf{[SF]} = \textbf{[NSF]}), and only the final sentence, \textit{"The caption is: 'There is no change.' or 'Straighten photo.'"}, is retained as the textual output \textbf{[T]}, with no \textbf{[LOC]} or \textbf{[SEG]} fields involved. Further details on the construction of in-context prompts are provided in Appendix~A.

By aligning sentence cues with structural fields (e.g., \textbf{[LOC]}, \textbf{[SEG]}), the model learns to perform structured reasoning and generate segmentation outputs as needed—without requiring explicit symbolic supervision.

\textbf{Structured Output Generation.}  
To generate the structured output \(C\), the model jointly encodes the input image pair, the contrast-enhanced visual features \(F_{\text{fused}}\), and an in-context prompt \(P\) that contains few-shot exemplars. The fused features are embedded as visual tokens, while the prompt provides linguistic priors to guide the output formatting. During decoding, the MLLM attends to both modalities to determine whether the observed change is segmentable and allocates semantic content to the corresponding output fields accordingly. Formally, the structured output \(C\) is defined as:
\begin{equation}
\label{eq:caption}
C = 
\begin{cases}
T, & \text{if } \textbf{[SF]} = \textbf{[NSF]}  \\
\textbf{[LOC]}_i,\ \textbf{[SEG]}_i,\ T, & \text{if } \textbf{[SF]} = \textbf{[YSF]} 
\end{cases}
\end{equation}
where \(T\) is the textual description, \(\textbf{[SF]} \in \{\textbf{[NSF]}, \textbf{[YSF]}\}\) is the segmentation flag predicted by CCC, and \(\textbf{[LOC]}_i\) and \(\textbf{[SEG]}_i\) represent the spatial region and pixel-wise segmentation mask for image \(I_i\), respectively.

This unified formulation allows the model to perform structured captioning through implicit supervision, without relying on manually designed tags or multi-head decoders. The output strictly adheres to the format defined in Equation~\ref{eq:caption}, enabling controllable generation that reflects both semantic meaning and spatial granularity.

To support this learning objective, we apply LoRA~\cite{hu2021lora} to the MLLM for efficient language adaptation, while the change visual fusion module is trained from scratch. The model is trained to autoregressively generate the structured caption \(C\) based on both visual and language context. The training loss is defined as:
\begin{equation}
\mathcal{L}_{\text{diff-text}} = - \sum_{t=1}^{T} \log P_{\theta}(c_t \mid c_{<t}, F_{\text{fused}}, P)
\end{equation}
\noindent where \(c_t\) is the \(t\)-th token in the output sequence, and \(\theta\) includes the parameters from both LoRA adapters and the fusion encoder.

\subsection{Chain-of-Change-Segmenting}
\label{sec:CCS}
As illustrated in the overall framework (Figure~\ref{fig:ccrc_overview}), the Chain-of-Change-Segmenting (CCS) module serves as the spatial localization stage within the CCRC framework. Its objective is to produce precise, pixel-level masks that delineate the semantic changes identified by the CCC module. The SAM encoder is omitted in the figure for clarity but included in implementation.

To reduce unnecessary computation and minimize interference from irrelevant background regions, CCS adopts a coarse-to-fine segmentation strategy conditioned on the predicted segmentation flag. 
Specifically, CCS is only activated when CCC predicts a segmentable change, i.e., \(\textbf{[SF]} = \textbf{[YSF]}\). Under this condition, CCC also provides a spatial prior \(\textbf{[LOC]}_i\) for each input image \(I_i\), which guides the segmentation process. The model first extracts image features \(f_{\text{seg}, i}\) via a frozen segmentation encoder \(F_{V2}\):
\begin{align}
    f_{\text{seg}, i} = F_{V2}(I_i),\quad i \in \{1, 2\}.
\end{align}

Next, a coarse \textit{Region Segmentation} mask \(\mathcal{D}_{i, t-1}\) is generated by feeding both the visual features and the spatial prompt \(\textbf{[LOC]}_i\) into the segmentation decoder \(D_{\text{sam}}\):
\begin{align}
\mathcal{D}_{i, t-1} = D_{\text{sam}}(f_{\text{seg}, i}, \textbf{[LOC]}_i).
\end{align}
This intermediate region mask \(\mathcal{D}_{i, t-1}\) is then encoded as a mask prompt using a dedicated encoder \(F_{\text{mask}}\), resulting in a segmentation prompt \(S_{i, t}\):
\begin{align}
S_{i, t} = F_{\text{mask}}(\mathcal{D}_{i, t-1}).
\end{align}

The \textit{Change-aware Token Refiner} (CATRefiner) module enhances visual-linguistic alignment by progressively refining the semantic embedding \([SEG]_{i,t}\) through multi-head change-aware attention (MHCA; see Section~\ref{sec:ccc}). To this end, we first concatenate visual features from two input images and extract a region-level change prototype using masked average pooling (MAP) over the union of their segmentation masks \(\mathcal{D}_{1, t-1} \oplus \mathcal{D}_{2, t-1}\). This prototype serves as a visual anchor to guide the update of \([SEG]_{i,t}\), enabling it to capture localized cross-image differences:
\begin{equation}
\hat{[SEG]}_{i, t} = F_{\text{ref}}\left(\beta([SEG]_{i, t}),\ f_{\text{seg},1} \oplus f_{\text{seg},2},\ \mathcal{D}_{1, t-1} \oplus \mathcal{D}_{2, t-1}\right),
\end{equation}
where \(\beta\) is an MLP projection layer, and \(F_{\text{ref}}(\cdot)\) is the refinement function of the CATRefiner:
\begin{equation}
F_{\text{ref}}(h,\ i,\ m) = h + \mathcal{A}_{\text{MHCA}}(h,\ \text{MAP}(i,\ m)).
\end{equation}
By injecting visual cues from both images into the reasoning chain, this module enables \([SEG]_{i,t}\) to focus more precisely on change-relevant regions, thereby improving semantic alignment and segmentation quality.

Finally, the model generates a refined \textit{Change Segmentation} mask \(\mathcal{D}_{i, t}\) by combining the visual features, segmentation prompt, and refined semantic embedding:
\begin{align}
\mathcal{D}_{i, t} = D_{\text{sam}}(f_{\text{seg}, i}, S_{i, t}, \hat{[SEG]}_{i, t}).
\end{align}

In this formulation, \(\mathcal{D}_{i, t-1}\) and \(\mathcal{D}_{i, t}\) respectively denote the intermediate Region Segmentation and the final Change Segmentation outputs for each image \(I_i\).

To ensure effective training of the segmentation process, we adopt a hybrid optimization strategy. The parameters of the projection layers and the mask decoder are set to be trainable, while the backbone components of the Segmentation Foundation Model (SAM)—including the image encoder and the prompt encoder—are kept frozen to retain their strong generalization capability.

During hierarchical segmentation, only the final-level segmentation mask is supervised, aligning the model's optimization with the ultimate localization objective. Like \cite{bao2024cores}, we employ a composite loss function that combines Dice loss and cross-entropy loss to capture both region-level overlap and pixel-wise classification accuracy:
\begin{align}
    \mathcal{L}_{\text{diff-seg}} = \lambda_d \cdot \mathcal{L}_{\text{Dice}}(S_T, S_{\text{GT}}) + \lambda_c \cdot \mathcal{L}_{\text{CE}}(S_T, S_{\text{GT}}),
\end{align}
where \( S_T \) denotes the predicted segmentation mask at the final stage, and \( S_{\text{GT}} \) is the corresponding ground truth mask. The balancing weights are empirically set to \( \lambda_d = 0.5 \) and \( \lambda_c = 2.0 \).

To jointly optimize both semantic understanding and spatial localization, the total training loss integrates the segmentation loss \(\mathcal{L}_{\text{seg}}\) and the captioning loss \(\mathcal{L}_{\text{text}}\) derived from the CCC chain:
\begin{align}
    \mathcal{L}_{\text{total}} = \lambda_R \cdot \mathcal{L}_{\text{diff-text}} + \lambda_S \cdot \mathcal{L}_{\text{diff-seg}},
\end{align}
where \( \lambda_R \) and \( \lambda_S \) balance the contributions of language and segmentation objectives, and are set to 0.5 by default.

\begin{table*}[t]
  \centering
  \caption{Comparison of ICCS performance on CLEVR-Change and Image-Editing-Request. B, M, C, and R denote BLEU-4, METEOR, CIDEr, and ROUGE; gIoU and cIoU assess segmentation quality.}
  \label{tab:test_comparison_clevr_ier}
  \resizebox{\textwidth}{!}{
    \begin{tabular}{l|cccccc|cccccc}
      \toprule
      \textbf{Method}
      & \multicolumn{6}{c|}{\textbf{CLEVR-Change}}
      & \multicolumn{6}{c}{\textbf{Image-Editing-Request}} \\
      \cmidrule(lr){2-7}\cmidrule(lr){8-13}
      & \textbf{B} & \textbf{M} & \textbf{C} & \textbf{R} & \textbf{gIoU} & \textbf{cIoU}
      & \textbf{B} & \textbf{M} & \textbf{C} & \textbf{R} & \textbf{gIoU} & \textbf{cIoU} \\
      \midrule
      X-Decoder(CVPR23)~\cite{zou2023generalized}
      & 45.3 & 28.2 & 110.3 & 64.2 & 24.3 & 17.5
      & 5.1 & 8.8 & 18.2 & 25.6 & 23.1 & 15.2 \\
      SEEM(NeurIPS23)~\cite{zou2023segment}
      & 48.2 & 30.8 & 115.8 & 65.3 & 25.6 & 19.3
      & 6.3 & 9.1 & 23.3 & 27.3 & 24.8 & 16.9 \\
      LISA (CVPR24)~\cite{lai2024lisa}
      & 54.2 & 38.1 & 128.9 & 72.3 & 50.2 & 44.0
      & 10.6 & 11.9 & 48.2 & 38.3 & 50.2 & 44.3 \\
      CoReS (ECCV24)~\cite{bao2024cores}
      & 55.1 & 38.4 & 130.7 & 72.6 & 52.1 & 45.3
      & 12.1 & 13.2 & 51.2 & 40.8 & 52.3 & 46.5 \\
      LISA++~\cite{yang2023lisa++}
      & 54.8 & 37.8 & 129.3 & 72.1 & 51.0 & 44.3
      & 9.9 & 12.4 & 50.9 & 39.7 & 51.9 & 45.8 \\
      SESAME (CVPR24)~\cite{wu2024see}
      & 56.3 & 39.1 & 132.3 & 75.3 & 51.2 & 45.5
      & 13.1 & 14.7 & 52.3 & 41.3 & 52.4 & 47.9 \\
      \midrule
      \textbf{Ours}
      & \textbf{58.3} & \textbf{42.5} & \textbf{139.4} & \textbf{77.5} & \textbf{54.4} & \textbf{48.7}
      & \textbf{15.3} & \textbf{17.5} & \textbf{55.8} & \textbf{44.2} & \textbf{54.7} & \textbf{49.1} \\
      \bottomrule
    \end{tabular}
  }
\end{table*}

\section{Experiments}
\subsection{Experimental setup}

\textbf{Implementation Details} Our model is built on LLaVA-v1.5-7B, which adopts Vicuna-7B as its language backbone, with a vision encoder (CLIP ViT-L-336px) and a two-layer MLP to enhance multimodal integration. The visual fusion module adopts a lightweight two-stage design, using one-layer Transformers (embed\_dim=256, num\_heads=8) for both single-image and pair-image encoding. The Change-aware Token Refiner shares the same configuration. We use the AdamW optimizer with weight decay and WarmupDecayLR scheduler (100 warmup iterations). For segmentation, we adopt the ViT-H variant of SAM. To ensure fair comparison, the same MLLM and SAM configuration is applied to LISA \cite{lai2024lisa}, CoRes~\cite{bao2024cores}, and our method.

\textbf{Datasets} This study uses the CLEVR-Change \cite{park2019robust} and Image-Editing-Request \cite{tan2019expressing} datasets, covering both synthetic and real-world changes. We annotate samples as either segmentable or non-segmentable. In the CLEVR-Change dataset, 39,803 samples are segmentable, and 39,803 are non-segmentable, totaling 79,606 samples. In the Image Editing Request dataset, 2,046 samples are segmentable, and 1,893 are non-segmentable, totaling 3,939 samples. Further annotation procedures and criteria are detailed in Appendix~B. 

\textbf{Evaluation} To evaluate Image Change Captioning and Segmentation, similar to previous work, we use metrics such as BLEU-4 (B4) \cite{papineni2002bleu}, METEOR (M) \cite{banerjee2005meteor}, CIDEr (C) \cite{vedantam2015cider}, ROUGE (R) \cite{chin2004rouge} for caption generation, with CIDEr as the core evaluation metric. For image segmentation, we refer to previous work and adopt two metrics: gIoU \cite{kazemzadeh2014referitgame} and cIoU \cite{mao2016generation}, with gIoU serving as the core evaluation metric.

\textbf{Handling Empty Targets}  Following Liu et al.~\cite{liu2023gres}, if a model correctly identifies an expression as referring to an empty target (i.e., no mask should be predicted), its gIoU is set to 1.0 and excluded from the cIoU computation. Conversely, if a mask is incorrectly generated, the gIoU is set to 0.0, and the predicted area contributes to the cIoU denominator, penalizing false positives. To identify empty targets, we apply model-specific rules: for \textbf{LISA} and \textbf{CoRes}, masks with fewer than 50 foreground pixels are considered empty. In contrast, \textbf{CCRC} uses a segmentation flag—if \textbf{[SF]} is \textbf{[NSF]}, the model deems the change non-segmentable and skips mask prediction. This consistent, interpretable strategy ensures fair evaluation across architectures.

To ensure fair evaluation, we reproduce X-Decoder~\cite{zou2023generalized}, SEEM~\cite{zou2023segment}, LISA~\cite{lai2024lisa}, LISA++ ~\cite{yang2023lisa++},  SESAME ~\cite{wu2024see} and CoReS~\cite{bao2024cores} based on their official implementations and released configurations. All models are evaluated within a unified framework.

\begin{table}[htbp]
    \centering
    \caption{Captioning performance under the ICC setting on CLEVR-Change. B, M, C, and R denote BLEU-4, METEOR, CIDEr, and ROUGE, respectively.}
    \label{tab:clevr_caption_only}
    \resizebox{0.48\textwidth}{!}{
        \begin{tabular}{lcccc}
            \toprule
            \textbf{Model} & \textbf{B} & \textbf{M} & \textbf{C} & \textbf{R} \\
            \midrule
            DUDA+Aux (CVPR21)\cite{hosseinzadeh2021image} & 51.2 & 37.7 & 115.4 & 70.5 \\
            SRDRL+AVS (ACL21)\cite{tu2021semantic}       & 54.9 & 40.2 & 122.2 & 73.1 \\
            VACC (ICCV21)\cite{kim2021agnostic}          & 52.4 & 37.5 & 114.2 & -    \\
            % MCCFormers-S (ICCV21)\cite{qiu2021describing} & 57.4 & 41.2 & 125.5 & -    \\ 
            IDC-PCL (AAAI22)\cite{yao2022image}          & 51.2 & 36.2 & 128.9 & 71.7 \\
            VARDTrans (TIP23)\cite{tu2023adaptive}       & 55.4 & 40.1 & 126.4 & 73.8 \\
            SCORER (ICCV23)\cite{tu2023self}             & 56.3 & 41.2 & 126.8 & 74.5 \\
            DIRL (ECCV24)\cite{tu2024distractors}            & 54.6 & 38.1 & 123.6 & 71.9 \\
            SMART (TPAMI24)\cite{tu2024smart}            & 56.1 & 40.8 & 127.0 & 74.2 \\
            DECIDER (AAAI25)\cite{zhong2025decider}           & 57.2 & 40.4 & 134.6 & 76.5 \\
            MCT-CCDiff (TIP25)\cite{hu2025mct}           & 58.0 & 41.1 & 134.8 & 76.3 \\
            FINER-MLLM (MM25)\cite{zhang2024differential}      & 55.6 & 36.6 & 137.2 & 72.2 \\
            \midrule
            \textbf{Ours}                                & \textbf{58.3} & \textbf{42.5} & \textbf{139.4} & \textbf{77.5} \\
            \bottomrule
        \end{tabular}
    }
\end{table}

\textbf{Results on CLEVR-Change} Table~\ref{tab:clevr_caption_only} presents the captioning results under the ICC setting. Our method achieves the best performance across most metrics, especially CIDEr (139.4) and ROUGE (77.5), outperforming state-of-the-art ICC models such as DECIDER~\cite{tu2024smart} and FINER-MLLM~\cite{zhang2024differential}, demonstrating superior semantic precision in describing image changes. Table~\ref{tab:test_comparison_clevr_ier} reports the full evaluation under the ICCS setting, which jointly assesses captioning and segmentation. All models follow a unified empty-target handling protocol: LISA \cite{lai2024lisa} and CoRes~\cite{bao2024cores} treat masks with fewer than 50 foreground pixels as empty, while CCRC uses the predicted flag \textbf{[SF]} and generates a mask only when \textbf{[SF]}=\textbf{[YSF]}. Our method consistently outperforms all baselines, achieving 54.4 gIoU and 48.7 cIoU, highlighting its robustness and the effectiveness of the dual-chain design.

\textbf{Results on Image-Editing-Request.}  
Table~\ref{tab:ier_caption_only} presents the captioning performance under the ICC setting. Our method achieves a CIDEr score of 55.8, outperforming FINER-MLLM by 2.3 points, with consistent improvements across BLEU, METEOR, and ROUGE. These results demonstrate the model's strong capability in generalizing to diverse and instruction-driven editing scenarios. Under the ICCS setting (Table~\ref{tab:test_comparison_clevr_ier}), our method also achieves superior segmentation performance, with a gIoU of 54.7 and a cIoU of 49.1, clearly outperforming LISA \cite{lai2024lisa} and CoRes~\cite{bao2024cores}. This confirms the effectiveness of the proposed dual-chain framework in jointly capturing fine-grained semantic changes and accurately localizing them, even under complex real-world image distributions.

\begin{table}[htbp]
    \centering
    \caption{Comparison of captioning performance with ICC models on Image-Editing-Request. B, M, C, and R denote BLEU-4, METEOR, CIDEr, and ROUGE, respectively.}
    \resizebox{0.45\textwidth}{!}{
    \begin{tabular}{lcccc}
        \toprule
        \textbf{Model} & \textbf{B} & \textbf{M} & \textbf{C} & \textbf{R} \\
        \midrule
        Rel-Att (ACL19)\cite{tan2019expressing}       & 6.7  & 12.8 & 26.4  & 37.4  \\
        % DUDA (ICCV19)\cite{park2019robust}            & 6.5  & 12.4 & 22.8  & 37.3  \\
        NCT (TMM23)\cite{tu2023neighborhood}          & 8.2  & 15.0 & 34.2  & 38.8  \\
        VARDTrans (TIP23)\cite{tu2023adaptive}        & 9.9  & 14.8 & 35.6  & 39.0  \\
        SCORER (ICCV23)\cite{tu2023self}              & 10.0 & 15.0 & 33.4  & 39.6  \\
        SMART (TPAMI24)\cite{tu2024smart}             & 10.5 & 15.2 & 37.8  & 39.1  \\
        DIRL (ECCV24)\cite{tu2024distractors}            & 10.9 & 15.0 & 34.1 & 41.0 \\
        DECIDER (AAAI25)\cite{zhong2025decider}       & 11.3 & 16.7 & 39.5  & 42.4  \\
        MCT-CCDiff (TIP25)\cite{hu2025mct}           &10.2 & 15.4 & 38.3 & 41.2\\
        FINER-MLLM (MM25)\cite{zhang2024differential}       & 14.1 & 15.9 & 53.5  & 40.4  \\
        \midrule
        \textbf{Ours}                                  & \textbf{15.3} & \textbf{17.5} & \textbf{55.8} & \textbf{44.2} \\
        \bottomrule
    \end{tabular}
    }
    \label{tab:ier_caption_only}
\end{table}

\begin{figure*}[t]
    \centering
    \includegraphics[width=0.92\textwidth]{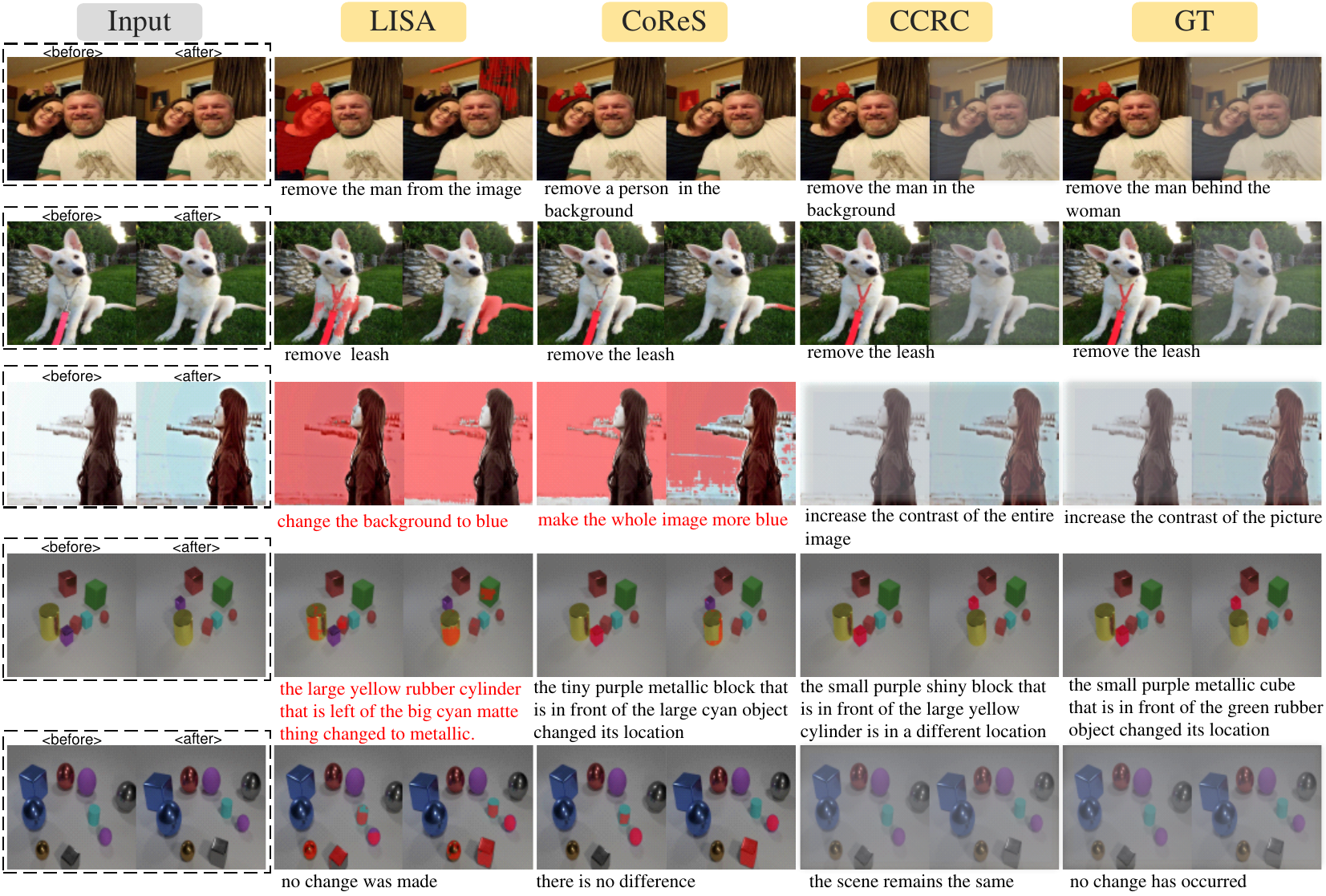}
    \caption{Qualitative results of LISA, CoReS, and CCRC on image change captioning and segmentation. Gray-shaded backgrounds denote non-segmentable changes (i.e., changes that cannot be localized for segmentation).}
    \label{fig:label_name}
\end{figure*}

\textbf{Qualitative Results:}
Figure~\ref{fig:label_name} compares the qualitative results of LISA, CoReS, and our CCRC framework on the ICCS task. The first three rows are from the Image-Editing-Request dataset, and the last two are from CLEVR-Change. Each example shows the generated captions and corresponding masks, with gray backgrounds indicating non-segmentable changes. The red-marked change captions highlight generation errors that are inconsistent with the ground-truth changes, including semantic inaccuracies and misaligned spatial references. CCRC achieves better performance in segmentability prediction, semantic-mask alignment, and spatial localization. It correctly identifies non-segmentable cases in real-world edits, avoids false masks, and produces more precise captions and tighter masks for object removals. In synthetic scenes, CCRC captures fine-grained changes and accurately models object relocations and structural variations, enabling more precise understanding. More qualitative results are provided in Appendix~D.

\subsection{Ablation Study}
To assess the contribution of each module in our framework, we conduct an incremental ablation study on the Image-Editing-Request dataset, as detailed in Table~\ref{tab:ier_ablation_incremental}. The base configuration adopts a LISA-style backbone with a LoRA-tuned MLLM and a directly trained segmentation encoder, augmented with a segmentability flag token \textbf{[SF]}. However, it lacks structured reasoning and spatial guidance mechanisms.
\begin{table}[ht]
    \centering
    \caption{Incremental ablation study on Image-Editing-Request. Components are progressively added to evaluate their contributions.}
    \label{tab:ier_ablation_incremental}
    \resizebox{0.48\textwidth}{!}{
    \begin{tabular}{lccccc}
        \toprule
        \textbf{Configuration} & \textbf{B} & \textbf{C} & \textbf{R} & \textbf{gIoU} & \textbf{cIoU} \\
        \midrule
        Base (LoRA + [SF])                     & 11.2 & 49.2 & 40.2 & 50.8 & 45.3 \\
        + CCC module                         & 12.7 & 51.5 & 40.6 & 51.2 & 46.1 \\
        + In-context prompting               & 12.9 & 51.7 & 40.9 &51.4 & 46.3 \\
        + Visual Fusion                      & 14.9 & 54.8 & 43.8 & 52.1 & 46.7 \\
        + CCS module                        & 15.1 &55.3  & 44.1 & 54.2 & 48.3 \\
        + CATRefiner (Full: CCRC)            & \textbf{15.3} & \textbf{55.8} & \textbf{44.2} & \textbf{54.7} & \textbf{49.1} \\
        \bottomrule
    \end{tabular}
    }
\end{table}

We begin by incorporating the Chain-of-Change-Captioning (CCC) module, which introduces semantic tokens \textbf{[LOC]} and \textbf{[SEG]} for coarse location and object-level reference. When \textbf{[SF]} is positive, their aggregated embedding provides weak supervision to the segmentation encoder, improving both captioning and segmentation (CIDEr↑ from 49.2 to 51.5; cIoU↑ from 45.3 to 46.1). Next, we add the In-context prompting strategy, offering format-aware language priors that enhance caption structural consistency and improve alignment (ROUGE↑ from 40.6 to 40.9; cIoU↑ to 46.3).

Subsequently, we enable the Visual Fusion module, which decouples intra-image and inter-image encoding. This module enhances contrastive modeling and cross-image difference capture, powered by our proposed Multi-Head Change-Aware Attention. This results in further improvements across both modalities (CIDEr↑ to 54.8; gIoU↑ to 52.1; cIoU↑ to 46.7). Introducing the Chain-of-Change-Segmenting (CCS) module yields substantial gains in spatial localization, thanks to its coarse-to-fine segmentation pipeline guided by CCC-derived priors (gIoU↑ from 52.1 to 54.2; cIoU↑ from 46.7 to 48.3). 

Finally, the full CCRC model, with the CATRefiner, achieves the best overall performance (CIDEr: 55.8; gIoU: 54.7; cIoU: 49.1), demonstrating the effectiveness of each module in enhancing semantic reasoning, structural guidance, and spatial precision. Further ablations on fusion layers and language models are in Appendix~C.

\section{Conclusion}
This paper presents Image Change Captioning and Segmentation (ICCS), a multimodal task that jointly models semantic change description and pixel-level localization. To address its dual objectives, we propose the CCRC framework, which decouples reasoning and segmentation through a dual-chain architecture. The Chain-of-Change-Captioning (CCC) module leverages Multi-Head Change-Aware Attention and in-context prompting for interpretable captioning, while the Chain-of-Change-Segmenting (CCS) module progressively refines localization based on CCC’s priors. We enhance both synthetic and real data with fine-grained segmentation to support thorough evaluation, and experiments show CCRC achieves state-of-the-art performance in both description and segmentation. Future work targets broader generalization in open-domain image changes.

% \ack We would like to thank the referees for their comments, which
% helped improve this paper considerably
% \bibliographystyle{plain}
\bibliography{mybibfile}

\end{document}